\documentclass[twocolumn,conference]{IEEEtran}
\usepackage[T1]{fontenc}
\usepackage[latin9]{inputenc}
\usepackage{xcolor}
\usepackage{amsbsy}
\usepackage{amssymb}
\usepackage{graphicx}

\makeatletter
\usepackage[caption=false,font=footnotesize]{subfig}
\usepackage{lipsum}
\usepackage{xcolor}

\usepackage{algorithm}
\usepackage{algpseudocode}
\usepackage{cite}
\usepackage{mathtools}

\makeatother

\begin{document}
\title{Dynamic Encoding and Decoding of Information for Split Learning in
Mobile-Edge Computing: Leveraging Information Bottleneck Theory}
\author{\IEEEauthorblockN{Omar~Alhussein\IEEEauthorrefmark{2}\IEEEauthorrefmark{1}, Moshi~Wei\IEEEauthorrefmark{3},
Arashmid~Akhavain\IEEEauthorrefmark{1}}\IEEEauthorblockA{\IEEEauthorrefmark{2}Department of Electrical Engineering and Computer
Science, Khalifa University, Abu Dhabi, UAE}\IEEEauthorblockA{\IEEEauthorrefmark{1}Advanced Networking Team, Huawei Ottawa Research
\& Development Centre, Ottawa, Canada}\IEEEauthorblockA{\IEEEauthorrefmark{3}Department of Computer Science, York University,
Toronto, Canada\\
omar.alhussein@ku.ac.ae,~arashmid.akhavain@huawei.com, moshiwei@yorku.ca}}

\maketitle

\begin{abstract}
Split learning is a privacy-preserving distributed learning paradigm
in which an ML model (e.g., a neural network) is split into two parts
(i.e., an encoder and a decoder). The encoder shares so-called latent
representation, rather than raw data, for model training. In mobile-edge
computing, network functions (such as traffic forecasting) can be
trained via split learning where an encoder resides in a user equipment
(UE) and a decoder resides in the edge network. Based on the data
processing inequality and the information bottleneck (IB) theory,
we present a new framework and training mechanism \textcolor{black}{to
enable a dynamic balancing of the transmission resource consumption
with the informativeness of the shared latent representations, which
directly impacts the predictive performance. The proposed training
mechanism offers an encoder-decoder neural network architecture featuring
multiple modes of complexity-relevance tradeoffs, enabling tunable
performance. The adaptability can accommodate varying real-time network
conditions and application requirements, potentially reducing operational
expenditure and enhancing network agility. As a proof of concept,
}we apply the training mechanism to a millimeter-wave (mmWave)-enabled
throughput prediction problem. We also offer new insights and highlight
some challenges related to recurrent neural networks from the perspective
of the IB theory. Interestingly, we find a compression phenomenon
across the temporal domain of the sequential model, in addition to
the compression phase that occurs with the number of training epochs.
\end{abstract}


\IEEEpeerreviewmaketitle{}

\section{Introduction}

\begingroup\renewcommand\thefootnote{\textsection}
\footnotetext{This work was conducted primarily at Huawei Ottawa Research \& Development Centre.}
\endgroup 

The dawn of the sixth-generation (6G) era is to bring forth a new
paradigm in communication networks, driven by the increasing demand
for ultra-fast, intelligent, and pervasive connectivity. As the digital
revolution permeates every aspect of our lives, there is an even greater
need to increase networking efficiency, enhance network automation,
and embed native intelligence. In 1949, Shannon and Weaver categorized
the problem of communications into three levels, namely (\emph{i})
the technical: transmission of symbols; (\emph{ii}) the semantic:
transmission of meaning; and (\emph{iii}) the effectiveness: effect
of semantic information exchange. While Shannon's communication model
considers the technical aspect only, there is a growing interest in
re-examining this fundamental consideration and incorporate semantics
to the 6G fabric.

Enabled by network softwarization and function virtualization, mobile-edge
computing rely on various network functions to optimize the network's
resources. Existing network functions, such as traffic forecasting,
traffic classification, packet scheduling, are increasingly being
implemented using experience-driven model-free machine learning (ML)
techniques. There is also an emergence of new functions related to
real-time sensing and analytics such as object detection and tracking.
Also, with 5G's service-based architecture, (over-the-top) application
functions can be hosted natively by service providers in core networks,
e.g., to enable virtual reality and close-proximity gaming. Given
the increasing adoption and diversity of network and application functions,
there is a need to optimize the communication and processing resources
to support and automate such network and application functions.

Hardware technology, as predicted by Moore's law, has been unable
to keep up with the exponential growth of computational and storage
requirements for modern ML models, resulting in a concerning widening
gap \cite{gholami2020ai_and_memory_wall}. New distributed learning
paradigms are needed to efficiently utilize available resources while
maintaining privacy and security. Split learning has emerged as a
distributed learning approach that offers a unique combination of
benefits for the wireless edge in terms of efficiency, privacy, and
flexibility \cite{poirot2019split,wu2023split}.\textcolor{brown}{}

In this paper, we identify and address a limitation with existing
split learning based predictive network and application functions;
the network substrate experiences time-varying usage behaviour and
traffic patterns mainly due to the time-varying and random behavior
of users. Moreover, network and application functions need to cater
towards diverse quality of service and predictive requirements. We
need to adapt the informativeness of encoded data in a dynamic manner
based on network conditions and application requirements to ensure
that the network remains communication-efficient and flexible under
a wide range of scenarios.   The information bottleneck (IB) method,
originally proposed by Tishby, Pereira, and Bialek, can provide a
new pair of lens for analyzing and improving deep learning models
\cite{tishby2000information,tishby2015deep}. The IB method attempts
to find the best tradeoff between the compression with regard to input
data and the preservation of relevant information needed for a specific
task based on the input\@. See Sections \ref{sec:IBT} and \ref{sec:literature}
for a continued discussion.

Building on this foundation, this paper introduces an adaptive neural
network encoding and decoding framework that adjusts the complexity-relevance
tradeoff in response to network conditions and application requirements.
We offer a training procedure that trains the neural network in a
cascaded (or tandem) fashion while connecting intermediate layers
from the encoder to the decoder to enable multiple modes of communication
with varying complexity-relevance modes. We apply the training procedure
to a mmWave 5G throughput prediction problem by utilizing the Lumos5G
dataset \cite{narayanan2020lumos5g}. The dataset captures throughput
as perceived by applications running on a user equipment (UE) along
with correlated features such as the longitude, latitude, and received
signal strength. 

Furthermore, this work is among the first to apply and investigate
the IB theory in the context of sequential (time-series) problems
and models. We provide new insights and identify important challenges
while partially addressing them. First, we find that incorporating
the temporal dimension of sequential models into the IB analysis is
essential for a comprehensive understanding of the training process.
Second, we find that estimating the mutual information (MI) can be
challenging due to the large hidden temporal states and sampling limitations.
Therefore, visualizing a 3-dimensional information plane that incorporates
the hidden temporal states and employing metrics such as the conditional
MI become vital to assess the redundancy of the temporal states, upon
which the number of states to be estimated can be reduced. A key finding
is that compression not only occurs as the training progresses (as
has been reported in the seminal IB works \cite{tishby2015deep,tishby2000information}),
but it also occurs across the temporal dimension of the sequential
model (i.e., across the hidden temporal states).

The rest of the paper is organized as follows. Section \ref{sec:IBT}
provides a brief background on IB. Section \ref{sec:literature} provides
an overview of relevant literature. Section \ref{sec:proposal} presents
the proposed dynamic framework and training mechanism. Section \ref{sec:mmwave}
discusses the use case of mmWave-enabled throughput prediction under
a split learning setup. In Section \ref{sec:numerical}, we conduct
experiments to confirm the viability of the proposed mechanism on
the Lumos5G dataset, along with a highlight of aforementioned challenges
and new insights. Section \ref{sec:conclusion} provides concluding
remarks.

\section{The Information Bottleneck Framework}

\label{sec:IBT}
\begin{figure}[htbp]
\begin{centering}
\textsf{\includegraphics[width=0.9\columnwidth]{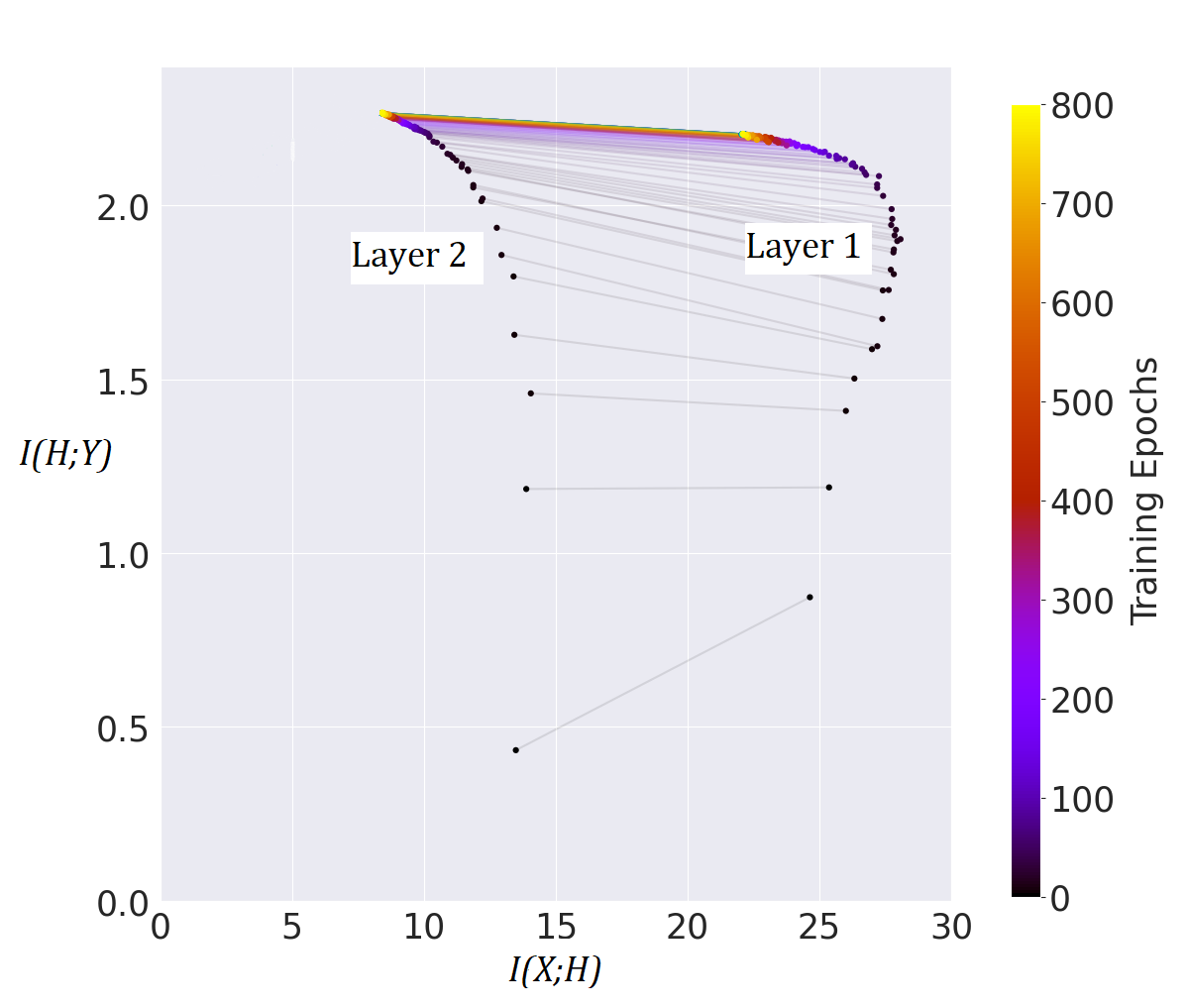}}
\par\end{centering}
\caption{Information plane showing the evolution of two layers as the training
progresses. Both layers exhibit a fitting phase followed by a compression
phase.}
\label{fig:info_plane_ex}
\end{figure}
Tishby and Zaslavsky suggested viewing each individual layer in a
neural network as a random variable \cite{tishby2015deep}. Therefore,
a neural network can be considered as a Markov chain of successive
representations. Using MI to gauge the flow of information can be
advantageous for two reasons. First, compared to statistical correlation,
MI is a more general measure of statistical dependence with a common
scale (e.g., bits or nats). Second, it is an invariant measure, i.e.,
\begin{equation}
I(X;Y)=I\big(\phi(X),\varphi(Y)\big),\label{eq:invariance}
\end{equation}
where $I(\cdot;\cdot)$ denotes the MI between two random variables,
and $\phi(\cdot)$ and $\varphi(\cdot)$ are deterministic functions.
The invariance property implies that MI can provide a unifying pair
of lens to probe and compare neural networks regardless of the architecture.
\textcolor{brown}{} IB provides a computational framework for finding
the optimal tradeoff between the compression of input data $X$ and
the preservation of information about target label $Y$ by minimizing
the Lagrangian,
\begin{equation}
\min_{p(H|X),p(H|T),p(H)}I(X;H)-\beta I(H;Y)\label{eq:IB_objective}
\end{equation}
where $\beta$ determines the level of relevant information captured
by neural network layer $H$. 

A useful graphical tool is the information plane which exhibits the
MI of a hidden layer with respect to target $Y$ versus the MI of
the hidden layer with respect to input $X$. \textcolor{black}{An
example of the information plane is shown in Fig. \ref{fig:info_plane_ex}
(obtained from our simulations). }Each curve in the information plane
corresponds to a layer. In the beginning, all layers gain information
with respect to both $X$ and $Y$ as the training progresses, which
is called the fitting phase. At some point in the training, $I(X;H)$
starts to reduce, exhibiting a compression phase. Saxe et al. provided
arguments and set of experiments that shows that compression occurs
only when the layers contain a double-saturating activation function
(such as tanh and sigmoid) \cite{saxe2019information}. Chelombiev
et al. showed that, by using adaptive and robust estimation techniques,
compression can occur without necessarily having double-edge saturation
in activation functions \cite{chelombiev2019adaptive}. This speaks
to the importance of deploying sensitive and robust estimation techniques.

\section{\textcolor{black}{Literature Survey}}

\label{sec:literature}Rate distortion theory, of Shannon and Kolmogorov,
characterizes the tradeoff between the signal representation and the
average distortion of the reconstructed signal \cite{cover1999elements}.
In the beginning of this century, Tishby et al. proposed IB as a generalized
distortion theory \cite{tishby2000information,slonim2002information}.
 In 2015 and 2017, Tishby et al. showed how the IB framework applies
to deep neural networks \cite{tishby2015deep,shwartz2017opening}.
Since then, there has been a great interest in the IB theory and its
applications \cite{hafez2019information,e22121408,git2022ib}. 

The literature can be classified into three categories. One category
attempts to further develop, analyze, and scrutinize the fundamentals
of the IB theory \cite{saxe2019information,kolchinsky2018caveats,achille2018emergence,piran2020dual,git2022ib}.
A second category applies IB principles to analyze and interpret the
inner workings of ML models, while a third category applies IB principles
and observations to improve deep learning-based algorithms and applications
\cite{hjelm2018learning}. The three categories overlap as further
development on the theory and deeper probing into the models can lead
to improving it.

To this paper's context, IB can fit in semantic communications \cite{strinati20216g,xie2021deep,qin2021semantic,gunduz2022beyond}
and wireless-edge learning \cite{park2019wireless,wang2020convergence}.
Notably, Beck et al. consider a semantic communication task such that
a message is transmitted while preserving the relevant meaning \cite{beck2023semantic}.
They cast the problem as an IB problem that allows messages to be
compressed while preserving the relevant information as possible.
Pezone et al. propose a goal-oriented system for edge-learning based
on the IB framework \cite{pezone2022goal}. They adapt $\beta$ to
optimize the tradeoff between the complexity and the relevance of
the encoded information in order to minimize the energy consumption
under delay constraints. Similarly, based on IB, Binucci et al. employ
convolutional encoders at the edge to compress relevant data before
being offloaded to an edge station \cite{gunduz2022beyond}.

\section{Adaptive Split Encoder-Decoder with feedback signals}

\label{sec:proposal}

\begin{figure}[tbph]
\begin{centering}
\textsf{\includegraphics[width=0.8\columnwidth]{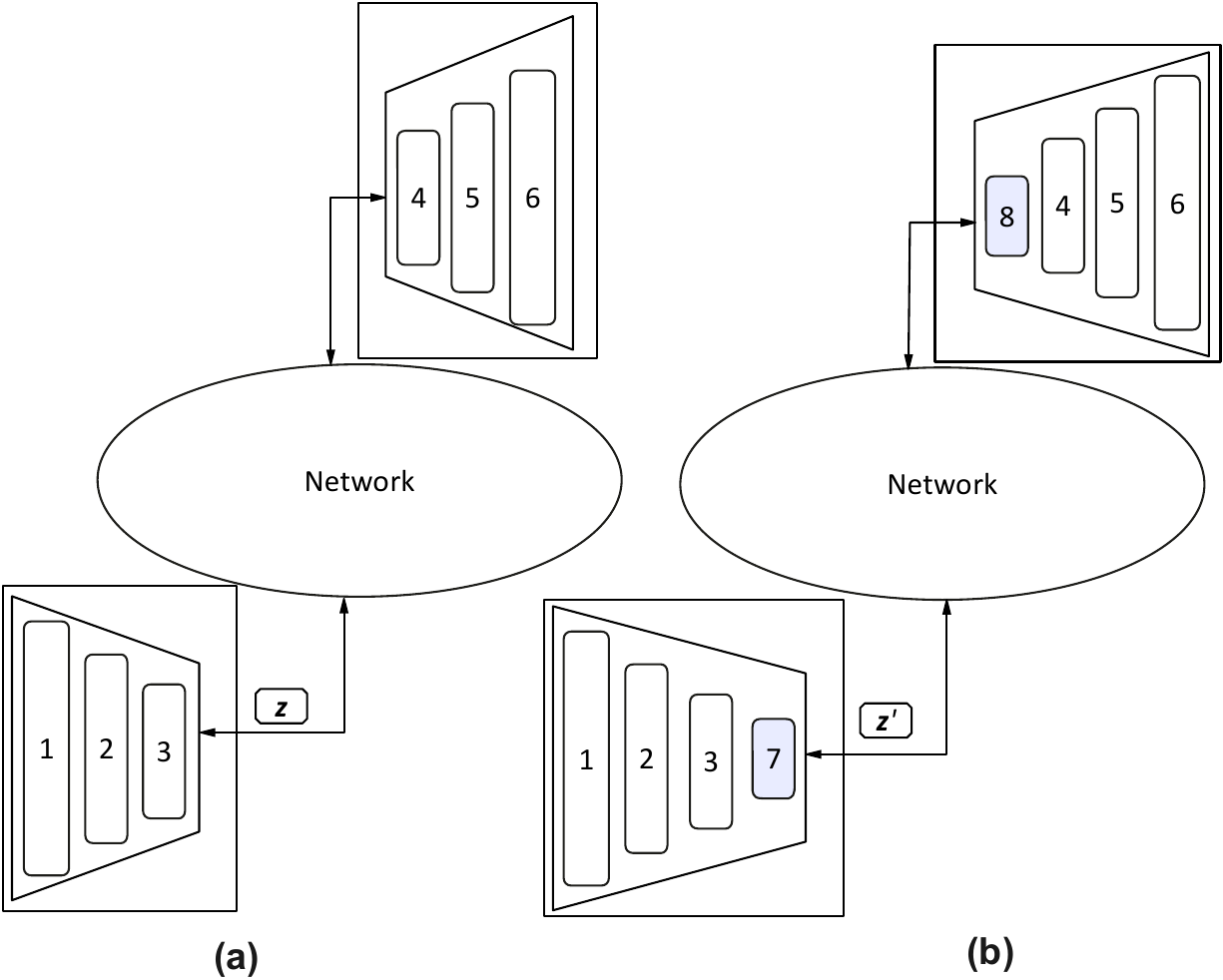}}
\par\end{centering}
\caption{Split encoder-decoder neural network where encoder produces two latent
codes; in (a) the encoder sends code $z$, whereas in (b) the encoder
augments one more layer and sends code $z'$.}
\label{fig:approach_step0}
\end{figure}
Network conditions and application requirements can vary over time
and between network slices. With the information plane in mind, the
intuition is to selectively choose which hidden layer\textquoteright s
output from the encoder to transmit to the decoder. Consider an optimized
neural network where the neural network hyper-parameters (i.e., number
of layers, number of nodes, learning rate) are optimized to provide
the best achievable predictive performance (e.g., through hyper-parameter
search). Denote the output of the optimized neural network encoder
by latent representation $\boldsymbol{z}$ (as shown in Fig. \ref{fig:approach_step0}(a)).
If we add a new bottleneck layer with output $\boldsymbol{z}'$ to
the trained encoder as shown in Fig. \ref{fig:approach_step0}(b),
following the data processing inequality and by construction, the
new encoder-decoder neural network is less optimal where adding an
additional bottleneck layer loses crucial information due to further
imposed compression. In Figs. \ref{fig:approach_step0}(a) and \ref{fig:approach_step0}(b),
$I(X;H_{3})\geq I(X;H_{7})$, entailing that code $\boldsymbol{z}'$
requires fewer bits than code $\boldsymbol{z}$ for encoding. It follows
that the decoder\textquoteright s predictive performance receiving
code $\boldsymbol{z}'$ is at most equal to or worse than the decoder\textquoteright s
performance receiving code $\boldsymbol{z}$. By combining both encoders
in Figs. \ref{fig:approach_step0}(a) and \ref{fig:approach_step0}(b),
we can get a dynamic and adaptive neural network encoder that can
vary the informativeness of latent representation depending on which
layer is selected as the transmitted bottleneck. In doing so, one
can train two decoders that receive representations $\boldsymbol{z}$
and $\boldsymbol{z}'$, respectively. A more compact solution that
results in one encoder and decoder is summarized in Algorithm \ref{alg:algorithm}.
Train a first encoder-decoder neural network (line 1). Freeze the
trained layers, add a new layer to the encoder and to the decoder,
respectively (lines 2-6), and retrain the overall network. Create
a (skip) connection between the output of the trained encoder and
the trained decoder (line 5). Finally, ensure that the new layers
results in a less optimal predictive performance after training (line
7), where the performance gap can be tuned by trial and error. \begin{algorithm} 
\caption{Cascaded training procedure to achieve two modes of complexity-relevance tradeoff}
\label{alg:algorithm} 
\begin{algorithmic}[1]
\State Encoder1, Decoder1 $\leftarrow$ Train([Encoder1, Decoder1])
\State Freeze(Encoder1, Decoder2)
\State NN2Encoder $\leftarrow$ [Encoder1 + new layer A]
\State NN2Decoder $\leftarrow$ [new layer B + Decoder1]
\Ensure Layer A output = layer B input
\State Connect Encoder1 and Decoder1
\State Encoder2, Decoder2 $\leftarrow$ Train([Encoder2,Decoder2])
\Ensure $I(Y;\mathrm{Decoder1Output}) \leq I(Y;\mathrm{Decoder2Output})$
\end{algorithmic} 
\end{algorithm}

\begin{figure}
\begin{centering}
\textsf{\includegraphics[width=0.8\columnwidth]{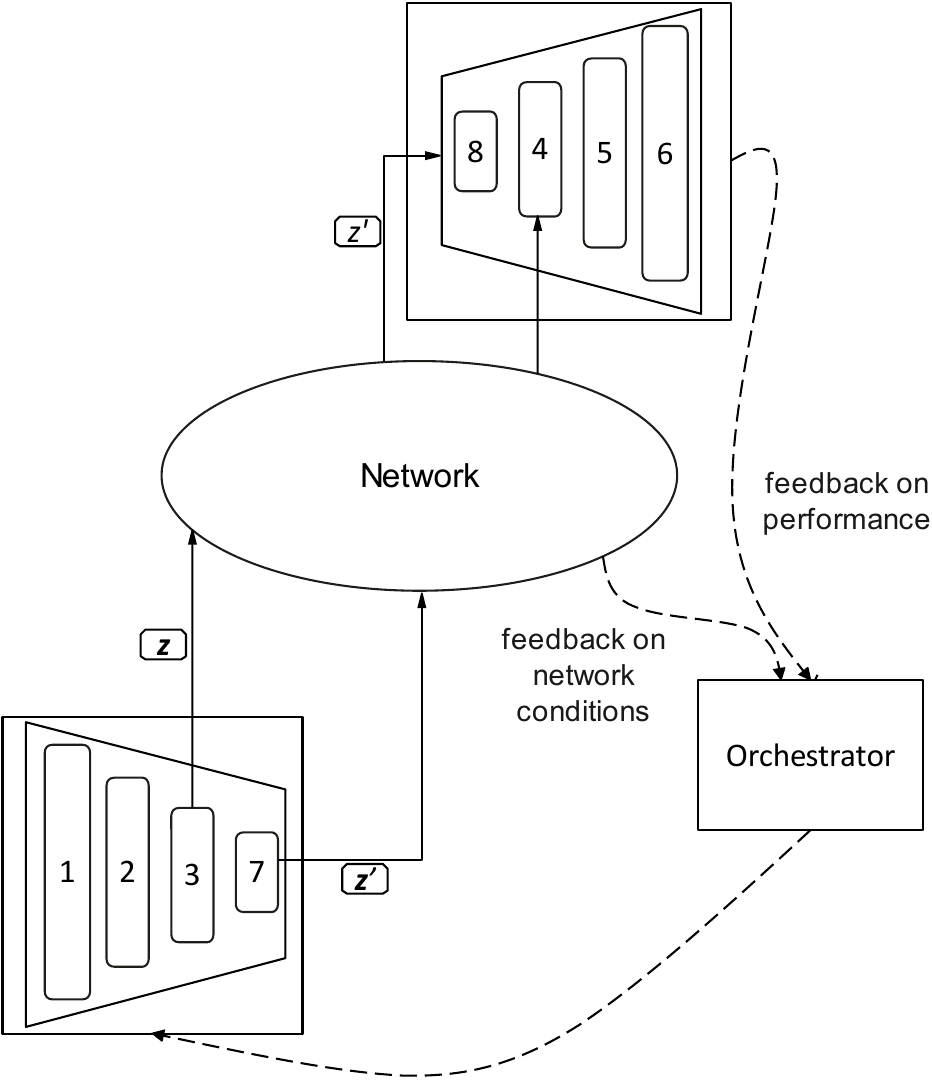}}
\par\end{centering}
\caption{Dynamic split neural network where the encoder can send either intermediate
latent code $z$ or less informative latent code $z'$. An orchestrator
monitors for network conditions and receives feedback about the decoder's
performance, and instructs the encoder to select which latent code
to transmit.}
\label{fig:general_method}
\end{figure}

Figure \ref{fig:general_method} depicts the general framework. Here,
an orchestrator can obtain key-performance indicators for each network
function from an oracle, and monitors the neural network's performance.
Based on the performance of the decoder and network conditions, the
orchestrator can instruct the network element containing the neural
network encoder to transmit either $\boldsymbol{z}$ and $\boldsymbol{z}'$.\textcolor{blue}{}
\textcolor{brown}{}
 \textcolor{red}{}\textcolor{brown}{}

\section{Use Case: mmWave Throughput Prediction}

\label{sec:mmwave} 
\begin{figure}[htbp]
\begin{centering}
\textsf{\includegraphics[width=0.4\columnwidth]{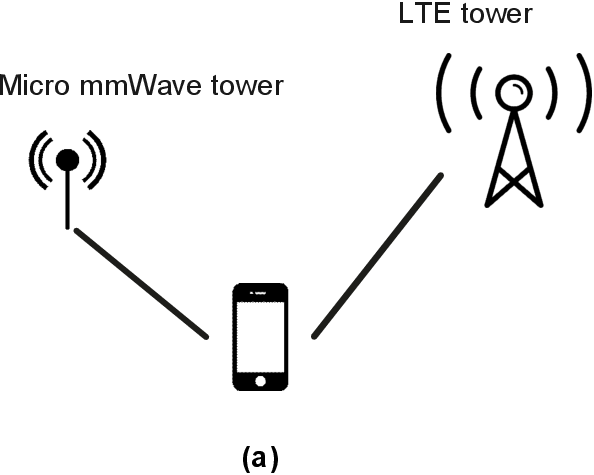}$\;$\includegraphics[width=0.4\columnwidth]{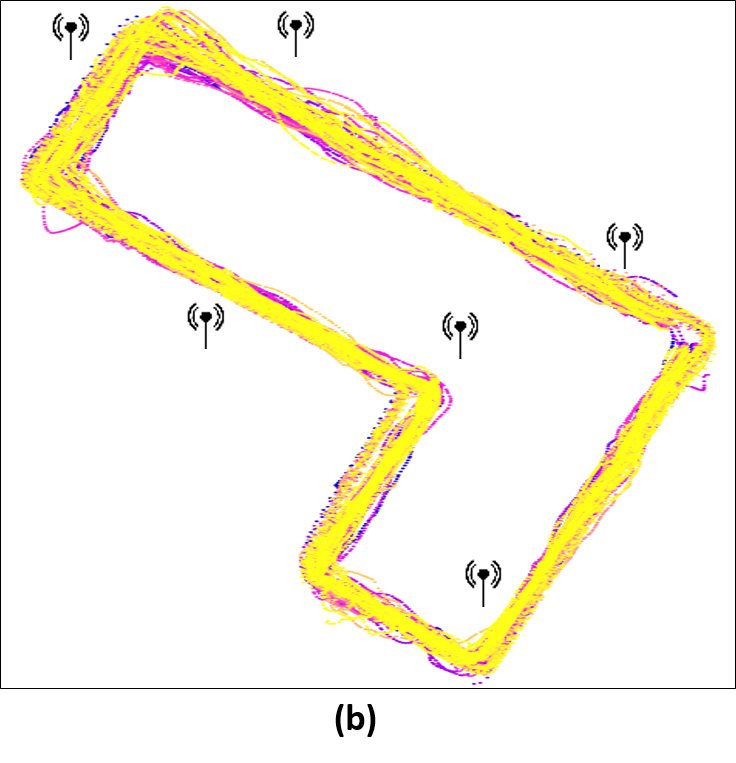}}
\par\end{centering}
\caption{(a) UE in dual-connectivity mode to a micro-mmWave based tower and
a macro-base station; (b) 1300 meter loop area in Minneapolis downtown
area based on the Lumos5G dataset \cite{narayanan2020lumos5g}. }
\label{fig:ue_sysmodel}
\end{figure}
 Consider a dual-connectivity setup where a UE is connected to a macro
base station (BS) and a mmWave-based micro BS, as shown in Fig. \ref{fig:ue_sysmodel}(a).
The user can get user-plane connections with high throughput via a
mmWave channel while having reliable control-plane (and user-plane)
connection to the macro BS. Since mmWave based beams are highly directed,
mobile users experience highly- and widely-varying throughput that
correlates with spatio-temporal features in the scene \cite{narayanan2020lumos5g}.
Many factors, such as position of the user, proximity of contending
users to each other, static and dynamic obstacles, user\textquoteright s
movement patterns and direction, have an effect on the perceived throughput.
The authors of \cite{narayanan2020lumos5g} captured some of these
features in the Lumos5G dataset which consists of 70,000 samples.
Each sample includes 11 features and an associated perceived throughput
by the UE. The features are longitude, latitude, moving speed, compass
direction, and six LTE and new-radio signal strength measurements,
c.f. \cite[Table 1]{narayanan2020lumos5g}. The samples are collected
along a 1300 meter loop area in Minneapolis downtown area as shown
in Fig. \ref{fig:ue_sysmodel}(b). Throughput prediction is useful
for over-the-top applications or network slices, such as video streaming,
real-time gaming, and virtual reality. For example, augmented reality
can cache/request information ahead of time depending on the predicted
throughput. Some slices/applications would require more accurate throughput
prediction performance compared to others.

In general, some of the useful spatio-temporal features can be observed
by the user equipment while others are can be obtained by the access
network through sensing or network measurements, e.g., distance of
the user to mm-wave BS, and dynamic/static obstacles in the vicinity.
\begin{figure}
\begin{centering}
\textsf{\includegraphics[width=1\columnwidth]{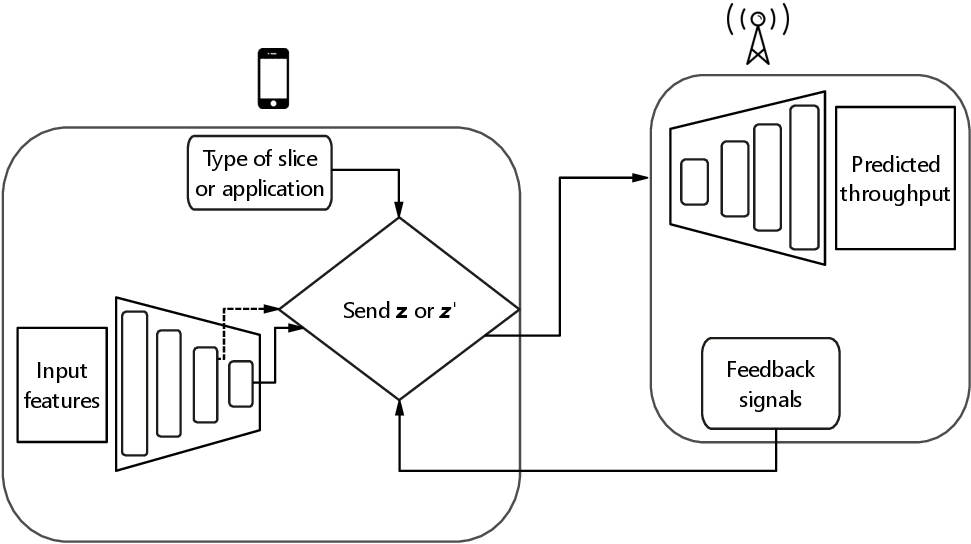}}
\par\end{centering}
\caption{Adaptive encoder-decoder network distributed between the UE and the
edge network. The encoder is trained/configured (using Algorithm \ref{alg:algorithm})
to send intermediate latent representations based on feedback signals.}
\label{fig:ue_adaptive}
\end{figure}
 Therefore, a split encoder-decoder model with encoders residing in
the UEs and a decoder residing in the access network is suitable to
learn experiences from the different users while minimizing feature
exchange and maintaining some level of privacy. \textcolor{brown}{}

The proposed dynamic split encoder-decoder framework can be utilized
for the mmWave throughput prediction problem. In Fig. \ref{fig:ue_adaptive},
a neural network encoder prompts the input features collected by the
UE (or received by the UE from other sources). Additionally, the encoder
can get performance requirements from the current application being
used. The access network can send back a feedback to the user of network
conditions, such as whether there is congestion on the allocated control-plane
band or whether computing hardware is congested in the mobile edge.
Depending on the feedback received and the type of slice/application
being used, the UE can decide whether to send the less-informative
output of the encoder (which in Fig. \ref{fig:ue_adaptive} is the
output of the $4$th layer) or the more-informative output from the
intermediate layer.

\begin{figure}[htbp]
\begin{centering}
\textsf{\includegraphics[width=1\columnwidth]{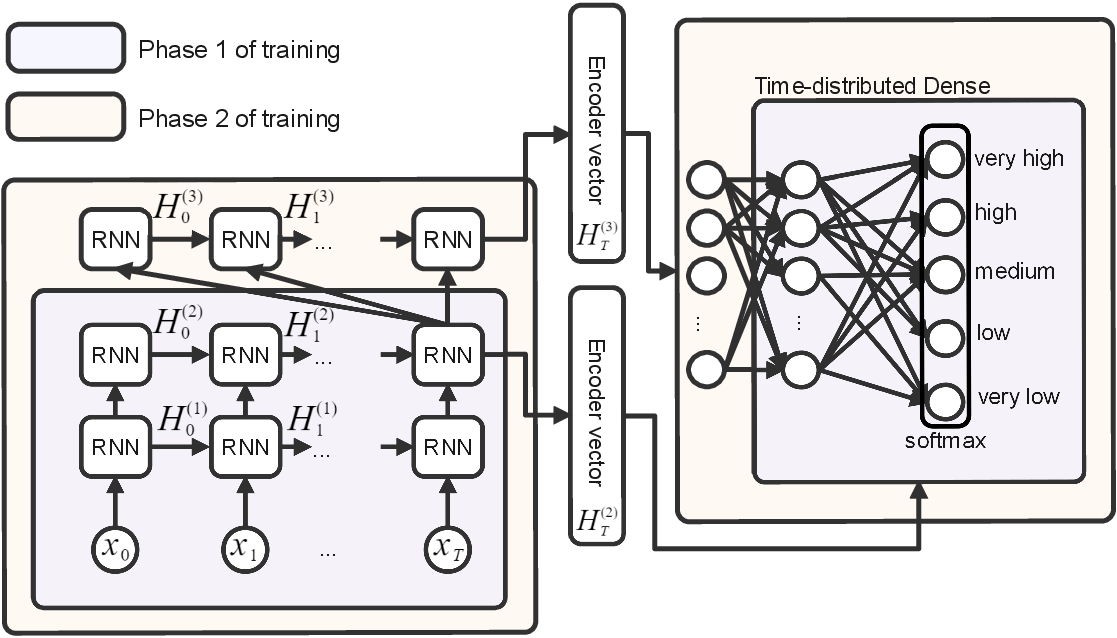}}
\par\end{centering}
\caption{Encoder-decoder LSTM-Dense model architecture trained in a cascaded
fashion.}
\label{fig:LSTM_timedistributed}
\end{figure}

Unlike some benchmark datasets used in the IB literature \cite{tishby2015deep,saxe2019information,tishby2000information},
here we have a time-series prediction problem. The input and output
time series can be modeled as random processes $X_{t},Y_{t}\in\mathbb{R}^{D}$,
where $t=1,\dots,T$ and $D$ correspond to the number of input features.
At each timestep $t$, $X_{t}$ has a corresponding latent state for
each layer $l$, $H_{t}^{(l)}$. We utilize a long-short term memory
(LSTM) network for the encoder and a time-distributed dense neural
network for the decoder, as shown in Fig. \ref{fig:LSTM_timedistributed}.
For the first phase of training, we train two LSTM layers for the
encoder followed by a time-distributed Dense neural network for the
decoder. Hyper-parameter search shows that two to three neural network
layers provide a good performance for the Lumos5G dataset. Following
Algorithm \ref{alg:algorithm}, we add a third LSTM layer at the encoder's
output and a time-distributed Dense layer at the decoder's input.
Then, we create an intermediate connection between the second layer
of the encoder and the second layer of the decoder. For each inference
query, the decoder receives either $H_{T}^{(2)}$ or $H_{T}^{(3)}$
\textit{but not both}. Here, note that latent codes transmitted from
$H^{(3)}$ goes through an extra layer of processing at the decoder
compared to latent codes transmitted from $H^{(2)}$.

\section{New Insights and Analysis}

\textcolor{black}{\label{sec:numerical}}In this section, we test
the main proposal on the Lumos5G dartaset where we construct the adaptive
LSTM-Dense encoder-decoder neural network following Algorithm \ref{alg:algorithm}
as shown in Fig. \ref{fig:LSTM_timedistributed}. The first two LSTM
layers have 128 cells, while the added layer after training has 32
cells. A common approach to estimate the mutual information using
the binning method which was employed in the original IB papers \cite{tishby2000information,tishby2015deep}.
The binning method quantizes the data into a number of bins, followed
by estimating the mutual information. Binning is sensitive on the
choice of the bin size and resultant boundaries, which affects the
accuracy and consistency of the mutual information estimate. We utilize
the kernel density estimator by Kolchinsky and others \cite{kolchinsky2017estimating,kolchinsky2019nonlinear,saxe2019information}
and the Gaussian-copula MI (GCMI) estimator by Ince and others \cite{ince2017statistical}
for $I(Y;T)$ and $I(X;T)$, respectively. The GCMI estimator is robust
to multidimensional variables and different marginal distributions.
Moreover, it can be extended to higher order quantities such as conditional
mutual information, which we strategize that it will be needed with
sequential or recurrent neural network models. In sequential models,
we not only care about the relation of a hidden layer with respect
to the input and the output, but we may need to quantify the effect
of a hidden state on a hidden state from a previous timestep, e.g.,
$\mathcal{I}(X;H_{j}|H_{j-1},H_{j-2})$.

Similar to \cite{narayanan2020lumos5g}, the input consists of $T=20$
timesteps, and the output of the decoder provides a classification
for 20 timesteps. The size of the testing set is set to $10\%$ of
the dataset, the learning rate is $10^{-2}$, and the batch size is
$256.$ 

\begin{figure}[htbp]
\begin{centering}
\textsf{\includegraphics[width=0.8\columnwidth]{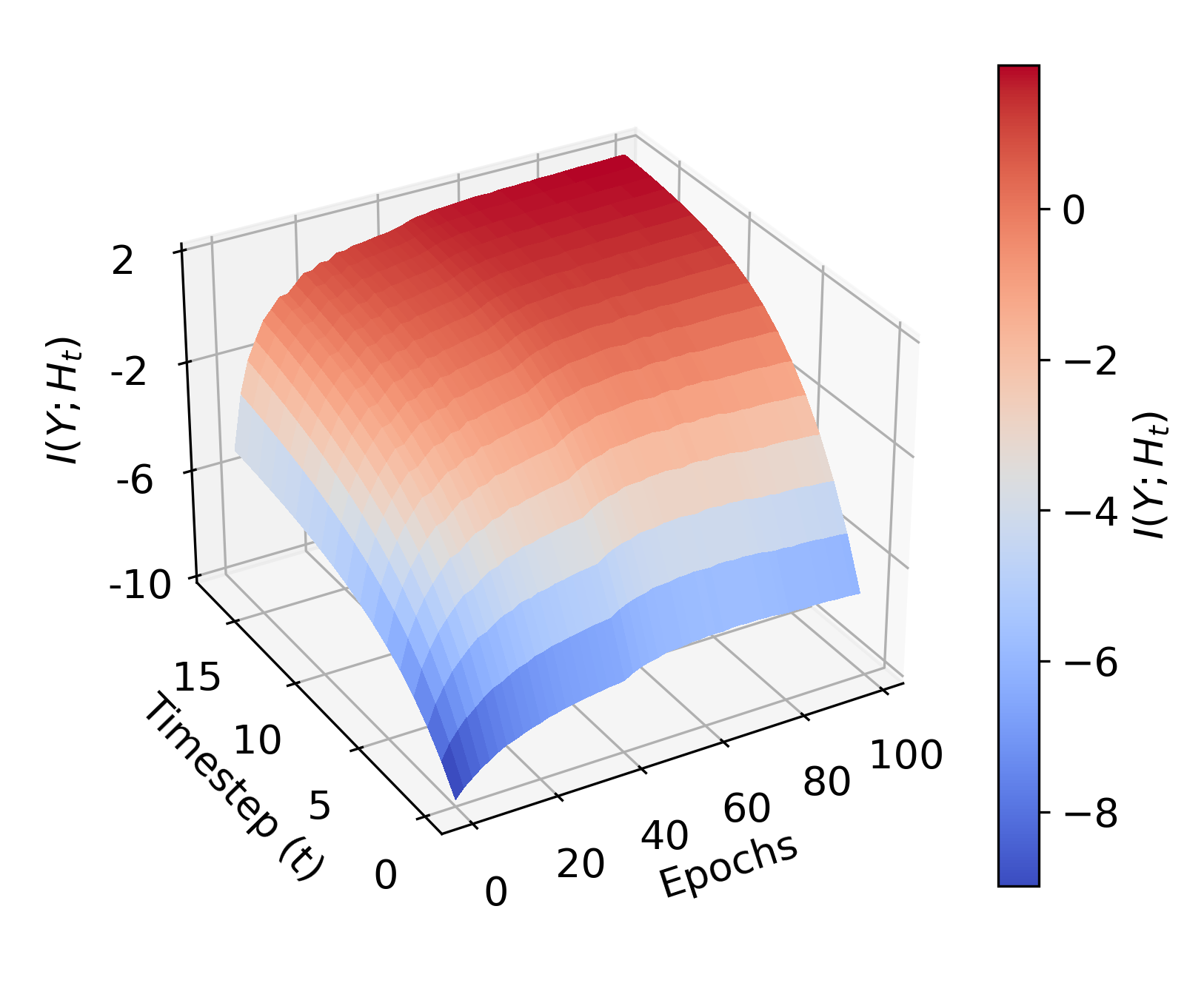}}
\par\end{centering}
\caption{3D Information curve: $I(H_{t};Y)$ with respect to timestep $t$
and number of training epochs. }
\label{fig:3d_infocurve}
\end{figure}

\begin{figure}[htbp]
\begin{centering}
\textsf{\includegraphics[width=0.9\columnwidth]{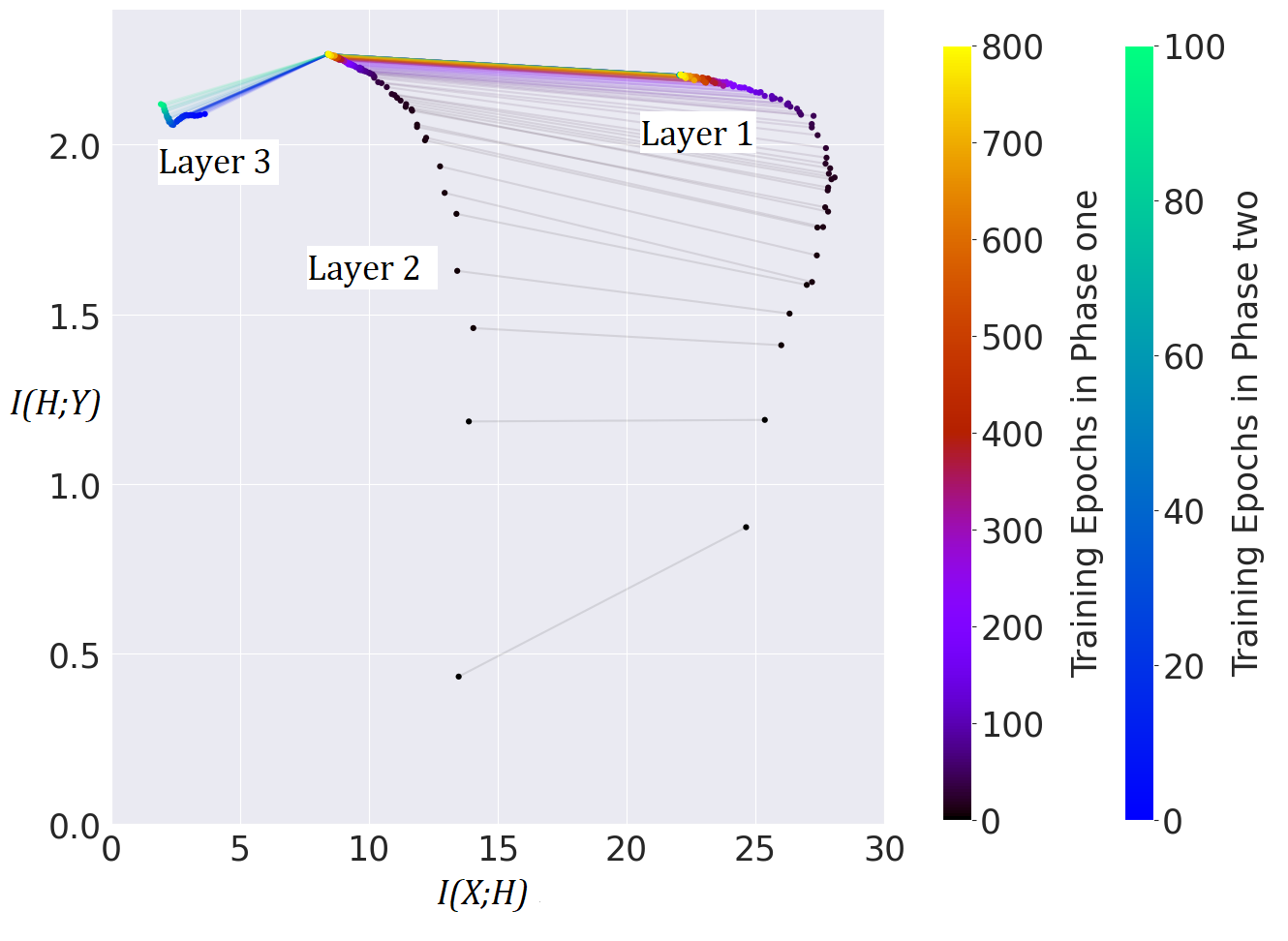}}
\par\end{centering}
\caption{The information plane showing the evolution of the neural network
layers' MI profiles, $I(H;Y)$ versus $I(H;X)$, with the number of
epochs. There are two training phases where starting from phase 2,
the encoder's neural network layers get frozen to train the added
layer.}
\label{fig:infoplane}
\end{figure}

From the perspective of the decoder, the first layer of the encoder's
LSTM network has a temporal state for each timestep, i.e., $H^{(1)}=[H_{1}^{(1)},H_{2}^{(2)},\dots,H_{T}^{(2)}],$
while the second layer conveys only the final temporal state, i.e.,
$H^{(2)}=[H_{T}^{(2)}]$. To view the information plane for the first
phase of training, we need to measure $I(\cdot;H^{(1)})$ and $I(\cdot;H^{(2)})$.
However, $H^{(1)}$ consists of 20 hidden temporal states, making
the estimation of the MI very difficult. To reduce the size of the
hidden state, we perform the following steps. First, Fig. \ref{fig:3d_infocurve}
shows a 3-dimensional plot of $I(H_{t}^{(1)};y_{\tau})$ versus timestep
$t$ and the number of epochs for $\tau=5$. 
\begin{figure}[htbp]
\begin{centering}
\textsf{\includegraphics[width=0.8\columnwidth]{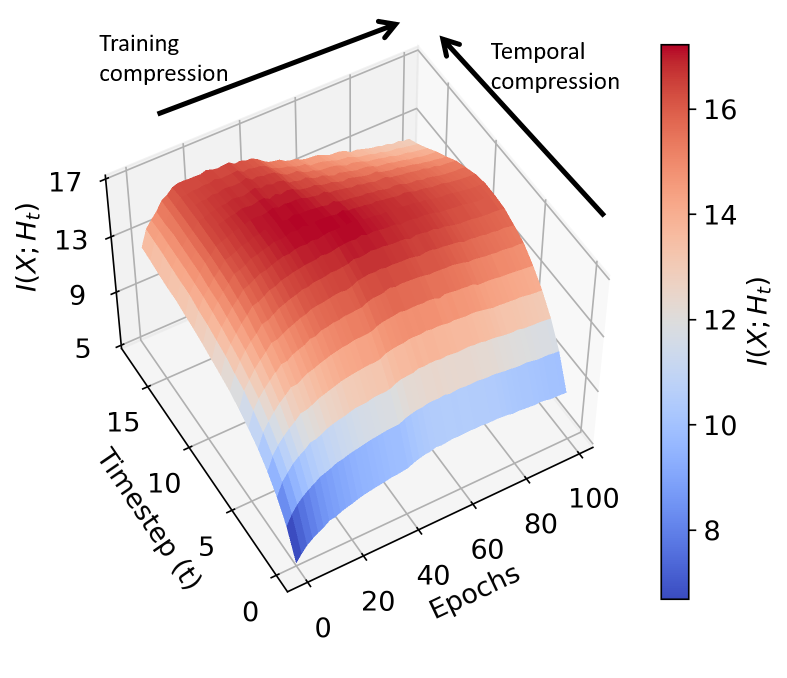}}
\par\end{centering}
\caption{3D Information curve: $I(H_{t};X)$ with respect to timestep $t$
and number of training epochs. The information curve shows compression
behavior with the training epochs and the hidden temporal states.}
\label{fig:3d_infocurve_x}
\end{figure}
 We observe that the last temporal state ($H_{T}^{(1)}$) contains
the largest amount of information, where $I(H_{t}^{(1)};y_{\tau})$
increases monotonically with $t$. Second, Fig. \ref{fig:3d_infocurve_x}
shows a 3-dimensional plot of $I(X_{1},\dots X_{t};H_{1},\dots,H_{t})$
versus timestep $t$ and the number of epochs. Interestingly, we find
that compression not only occurs as the training progresses (i.e.,
with the number of epochs), but it also occurs across the hidden temporal
states\textcolor{black}{, which to our knowledge has not been reported
before in the open literature.} The observed trends in Figs. \ref{fig:3d_infocurve}
and \ref{fig:3d_infocurve_x} were found to be consistent for other
values of $\tau$. The trends imply that there can be strong redundancy
across the hidden temporal states and that the last few temporal states
can be sufficient to represent all the hidden states. Thanks to the
flexibility of the GCMI estimation technique \cite{ince2017statistical},
we can also measure the conditional MI to gauge the amount of redundancy
contained in the temporal hidden states. We find that $I(x_{1},\dots,x_{T};H_{T}^{(1)}|H_{T-1}^{(1)})=14.24$
bits, $I(x_{1},\dots,x_{T};H_{T}^{(1)}|H_{T-1}^{(1)},H_{T-2}^{(1)})=3.23$
bits, and $I(x_{1},\dots,x_{T};H_{T}^{(1)}|H_{T-1}^{(1)},H_{T-2}^{(1)},H_{T-3}^{(1)})=2.37$
bits. The conditional MI keeps decreasing as we condition on earlier
hidden states. This implies that useful information are sufficiently
represented in the last few temporal states. Therefore, for the information
plane in Fig. \ref{fig:infoplane}, we consider 
\begin{equation}
H^{(1)}\approx[H_{T}^{(1)},H_{T-1}^{(1)},H_{T-2}^{(1)},H_{T-3}^{(1)}].\label{eq:approx_fromGCMI}
\end{equation}

Fig. \ref{fig:infoplane} shows the information plane for both training
phases. In the first training phase, the two layers converge to $I(H;y_{t})=2.3$
bits approximately, and $I(H;X_{1},\dots,X_{T})$ of 17 bits for the
first layer and 9 bits for the second layer. Following Algorithm \ref{alg:algorithm},
in the second phase, we freeze the trained neural network layers,
add a new bottleneck layer to the encoder, and re-train the neural
network.
 In the second training phase, the added layer converges to $I(H_{T}^{(3)};x_{1},\dots,x_{T})=2.5$
bits and $I(H_{T}^{(3)};y_{t})=2.25$ bits. Note how adding the new
bottleneck layer reduces $I(H_{T}^{(3)};y_{t})$ and correspondingly
the prediction performance, which is intended by design. Using the
framework explained in Sections \ref{sec:proposal} and \ref{sec:mmwave},
we can adaptively switch between the outputs of encoder layer 2 and
encoder layer 3, respectively.

The impact of a reduction in the target MI and the predictive performance
of a network function on the overall network performance remains unclear.
We suggest framing this issue as an optimization or search problem,
aiming to minimize signaling overhead or feature exchange under specific
conditions without significantly compromising network's performance.
Additionally, this study highlights the challenge of estimating the
MI in sequential models.

\section{Conclusions}

\label{sec:conclusion}Based on the IB theory and the data processing
inequality, this paper presents a dynamic framework and a training
mechanism to tune the informativeness of the shared latent representation
for split-learning based network functions. This dynamic tunability
provides flexibility to address varying network conditions and application
requirements. We apply the training mechanism to a mmWave throughput
prediction problem using the Lumos5G dataset as a proof of concept.
This paper highlights (\emph{i}) the importance of incorporating the
temporal domain into the IB analysis, and (\emph{ii}) the challenge
of estimating the information plane in sequential models which can
contain a large number of hidden temporal states. Interestingly, we
also observe a compression phenomena that occurs across the temporal
domain in sequential models.

\bibliographystyle{IEEEtran}
\bibliography{references}

\begin{IEEEbiography}{}
\end{IEEEbiography}

\end{document}